\documentclass[%
 aip,
 amsmath,amssymb,
 reprint,%
]{revtex4-1}
\usepackage[utf8]{inputenc}
\usepackage{comment}
\usepackage{bbm}
\usepackage[normalem]{ulem}
\usepackage{amsmath,bm}
\usepackage{amssymb}
\usepackage{graphicx}
\usepackage{booktabs}
\usepackage{xcolor}
\usepackage{xr}
\usepackage{url}
\usepackage{hyperref}
\usepackage[margin=1in]{geometry}
\begin{document}

\title{A survey of probabilistic generative frameworks for molecular simulations}

\author{Richard John}
\affiliation{Department of Physics and Institute for Physical Science and Technology, University of Maryland, College Park, MD, 20742, USA}
\author{Lukas Herron}
\affiliation{Biophysics Program and Institute for Physical Science and Technology, University of Maryland, College Park, MD, 20742, USA}
\affiliation{University of Maryland Institute for Health Computing, Bethesda 20852, USA.}
\author{Pratyush Tiwary}
\affiliation{University of Maryland Institute for Health Computing, Bethesda 20852, USA.}
\affiliation{Department of Chemistry and Biochemistry and Institute for Physical Science and Technology, University of Maryland, College Park, MD, 20742, USA.}

\begin{abstract}
Generative artificial intelligence is now a widely used tool in molecular science. Despite the popularity of probabilistic generative models, numerical experiments benchmarking their performance on molecular data are lacking. In this work, we introduce and explain several classes of generative models, broadly sorted into two categories: flow-based models and diffusion models. We select three representative models: Neural Spline Flows, Conditional Flow Matching, and Denoising Diffusion Probabilistic Models, and examine their accuracy, computational cost, and generation speed across datasets with tunable dimensionality, complexity, and modal asymmetry. Our findings are varied, with no one framework being the best for all purposes. In a nutshell, (i) Neural Spline Flows do best at capturing mode asymmetry present in low-dimensional data, (ii) Conditional Flow Matching outperforms other models for high-dimensional data with low complexity, and (iii) Denoising Diffusion Probabilistic Models appears the best for low-dimensional data with high complexity. Our datasets include a Gaussian mixture model and the dihedral torsion angle distribution of the Aib\textsubscript{9} peptide, generated via a molecular dynamics simulation. We hope our taxonomy of probabilistic generative frameworks and numerical results may guide model selection for a wide range of molecular tasks.
\end{abstract}

\maketitle

\section{Introduction}

In recent years, generative artificial intelligence (AI) has demonstrated a remarkable capacity to produce convincing images, text, audio, and video\cite{latent-diffusion,chatgpt,video-diffusion}. The domain of applicability of generative AI has recently extended to the molecular sciences \cite{tiwary2024generative}, where generative AI has demonstrated the ability to predict protein tertiary structure from amino acid sequence \cite{distributional-graphormer,alphafold-2,alphafold-3,RF-AA}, protein-ligand complex tertiary structure from chemical identity \cite{NeuralPLexer, diffdock} and the temperature dependence of the equilibrium distribution of solvated molecular systems \cite{BoltzmannGenerators, ddpm-wang-2022, thermomaps-2023}. While these methods differ in many aspects, all generative models share the common goal of sampling from an unknown underlying probability distribution based on an empirical dataset.

While there are many classes of generative models, recently, probabilistic generative models have seen widespread usage. These models represent a framework that broadly encompasses flow-based \cite{RealNVP,flow-matching-ot} and diffusion models\cite{score-based-generative}. The probabilistic generative framework explicitly seeks to directly model the data distribution through a series of invertible transformations (in flow-based models) or by iteratively refining noisy samples back into data space (in diffusion models), providing a flexible method for generating new data points that obey the underlying distribution of the observed data.

There is now a range of probabilistic generative models for use in different domains. Arguably, the most popular ones include Neural spline Flows\cite{neural-splines} (NS) models, Conditional Flow Matching\cite{CFM} (CFM) models, and Denoising Diffusion Probabilistic Models\cite{ddpm} (DDPM). All of these have already been used for exciting and novel applications, including sound field reconstruction \cite{2024soundfield} (NS), zero-shot text-to-speech synthesis \cite{kim2023pflow} (CFM), and medical image segmentation \cite{Guo2023} (DDPM), demonstrating the utility of probabilistic generative models across different modalities. We describe these methods in Section \ref{sec:gen-frameworks}. However, the scientific literature in this field lacks a systematic comparison of these methods for benchmark problems with tunable complexities that could establish the conditions under which one particular framework out of NS, CFM, and DDPM might be advantageous. This is particularly true for applications to molecular systems. In this work, we address this gap by carefully applying NS, CFM, and DDPM to different benchmark systems. We realize that the field is moving extremely quickly, with new variants of flow and diffusion methods appearing regularly. In this vein, we expect that the datasets used here will serve as useful benchmarks for these new methods.

Our systems include a Gaussian mixture model (GMM) and an explicit water molecular dynamics trajectory for the Aib\textsubscript{9} peptide \cite{shams-aib9, botan-peptide}, where we collect information on the $\{ \Phi,\Psi \}_i$ dihedral angles for all 9 residues. For the Gaussian mixture model dataset, we are interested in how generative model accuracy scales with data dimensionality and with training dataset size and which model best estimates probability density differences between asymmetric modes in the training dataset. We also measure sample generation speed and model network size as data dimensionality varies. For Aib\textsubscript{9}, we are interested in model performance on molecular dynamics data at varying levels of complexity, which we tune by looking at different residues within the peptide. We also examine model accuracy in the low training data limit for the Aib\textsubscript{9} dataset. Overall, our findings are:

\begin{itemize}
    \item NS exhibits superior performance estimating probability density differences. However, NS accuracy decreases for high-dimensional data.
    \item CFM displays the highest accuracy at high dimensionality but diminished performance in the presence of complex, multiple modes.
    \item DDPM most accurately models the complex, multimodal Aib\textsubscript{9} dihedral angle distribution. However, DDPM is less accurate than other methods at high data dimensionality.
\end{itemize}

\section{Theoretical background}

The microscopic probabilities of configurations of a system comprising coordinates $\mathbf{x}\in \mathbb{R}^d$ are described by the Boltzmann distribution
\begin{equation}
    \label{eq:boltzmann}
    p(\mathbf{x}) = \frac{e^{-\beta U(\mathbf{x})}}{Z(\beta)} \; \mathrm{with} \; Z(\beta) = \int e^{-\beta U(\mathbf{x})} d\mathbf{x},
\end{equation}
where $\beta$ is the inverse temperature, $U(\mathbf{x})$ is the energy and $Z(\beta)$ is the normalizing constant of $p(\mathbf{x})$, also known as the partition function. Upon first impression, Eq. \ref{eq:boltzmann} may seem trivial, but the relationship between $Z(\beta)$ and $p(\mathbf{x})$ is subtle. The moments of $U(\mathbf{x})$ with respect to $p(\mathbf{x})$ are generated by the derivatives of $\ln Z(\beta)$, i.e.
\begin{equation}
    \label{eq:pfunc-derivative}
    \frac{\partial \ln Z(\beta)}{\partial \beta} = -\langle U (\mathbf{x})\rangle_{p(\mathbf{x})}
\end{equation}
where the angular brackets denote ensemble averaging with respect to $p(\mathbf{x})$.

Since the underlying structure of $p(\mathbf{x})$ at temperature $\beta$ -- a potentially complex, high-dimensional probability distribution -- is encoded in changes in $Z(\beta)$ -- a scalar-to-scalar function -- significant effort has been devoted to developing computational strategies to estimate changes in partition functions, or equivalently free energy differences, where the free energy is defined as:
\begin{equation}
    F(\beta) = -\beta^{-1} \ln Z(\beta).
\end{equation}

The derivative in Eq. \ref{eq:pfunc-derivative} indicates that the partition function is a relational quantity -- that is, changes in the partition function are thermodynamically meaningful rather than the value of the function itself. Likewise, as a quantity derived from the partition function, free energy differences are typically of interest rather than absolute free energies.

Free energy differences are evaluated between a {\it target} state described by Boltzmann distribution $p(\mathbf{x})$ and a {\it reference} state with distribution $q(\mathbf{x})$. Assuming that the target and reference states share the same temperature (say $\beta=1$), one may express the free energy difference as a ratio of partition functions:
\begin{equation}
    \label{eq:fep-identity}
    \Delta F_{pq} = - \ln \frac{Z_q}{Z_p}.
\end{equation}

Computing the free energy difference in this fashion requires evaluating the Boltzmann weights (the integrand of Eq. \ref{eq:boltzmann}) using the states' energy functions. In cases where the energy function is unknown, e.g., if the configuration space comprises collective variables, then the Kullback-Leibler (KL) divergence provides an upper bound on $\Delta F_{pq}$. The KL divergence between $p(\mathbf{x})$ and $q(\mathbf{x})$ is defined as
\begin{equation} 
\label{eq:KLD} 
D_{\mathrm{KL}}(p||q) = \int p(\mathbf{x}) \log \frac{p(\mathbf{x})}{q(\mathbf{x})} d\mathbf{x}
\end{equation} 
and is a measurement of similarity between $p(\mathbf{x})$ and $q(\mathbf{x})$. More specifically, if $D_{KL}(p||q)=0$, then $\Delta F_{pq}=0$ and distributions $p$ and $q$ are identical.

\section{Probabilistic Generative Frameworks}
\label{sec:gen-frameworks}

Probabilistic generative models yield samples from an intractable {\it target} distribution $p(\mathbf{x})$ by transforming a simpler {\it prior} distribution $q(\mathbf{x}')$ into the target distribution. The change of measure identity underlies probabilistic generative models: it states that the change of probability as a result of an invertible coordinate transformation $\mathcal{M}: \mathbf{x}' \rightarrow \mathbf{x}$ is
\begin{equation}
    \label{eq:change-of-measure}
    p(\mathbf{x}) = \frac{q(\mathbf{x}')}{|J_\mathcal{M}(\mathbf{x}')|},
\end{equation}
where $J_\mathcal{M}$ is the Jacobian of $\mathcal{M}$. 

Generally, the objective of a probabilistic generative model is to find an $\mathcal{M}$ that minimizes the free energy difference between a set of empirical samples $\mathcal{D}$, and $\mathcal{M}$ applied to $q(\mathbf{x}')$. 

Once obtained the map is demonstrably useful, e.g. for accelerating the convergence of free energy estimates via targeted free energy perturbation\cite{tfep-2002,bijective-maps-2009,bg-2019,tfep-2020,tfep-2021,lrex-2022,thermomaps-2023}. However, there are several approaches to optimizing $\mathcal{M}$, each with advantages and disadvantages. In Sections \ref{sec:normalizing-flows}-\ref{sec:sbm} we summarize neural network-based approaches to optimizing $\mathcal{M}$.

\subsection{Normalizable Architectures}
\label{sec:normalizing-flows}

Normalizing flows are a class of probabilistic generative models wherein a neural network defines an invertible map $f_\theta$ with change in probability
\begin{equation} 
\label{eq:nf} 
\log p(\mathbf{x}) = \log q(f_\theta(\mathbf{x})) - \log \left|  J_{f_\theta}(\mathbf{x}) \right|.
\end{equation} 

The network is optimized by maximizing (minimizing) the likelihood (free energy) of $\mathcal{D}$ under the right-hand side of Eq. \ref{eq:nf}. Once learned, a sample from $p(\mathbf{x})$ may be obtained by first sampling $q(\mathbf{x}')$ and then applying the inverse map $f_\theta^{-1}$. The change in probability may be computed by evaluating the Jacobian determinant $J_{f_\theta}$. 

Computationally evaluating the Jacobian determinant is expensive; in the general case, the complexity scales cubically with the dimension $d$, but imposing additional structure on the transformation may simplify the calculation. For example, the complexity for a triangular Jacobian is linear in $d$. Normalizable architectures impose additional structure on the operations the network performs in order to simplify the determinant calculation, such as using layers that alternately produce upper- and lower-triangular Jacobians\cite{RealNVP}. In practice, however, the additional structure limits the expressivity of the network. Since the introduction of the framework, efforts have focused on balancing expressivity and computational feasibility\cite{glow,neural-splines,masked-autoregressive-flows,stochastic-normalizing-flows}.

\subsection{Neural Ordinary Differential Equations}
\label{sec:neural-ode}
Neural Ordinary Differential Equations (ODEs) \cite{CNF} form the basis of diffusion and flow matching models. Neural ODEs use neural networks to model the solution of a differential equation:
\begin{equation}
    \label{eq:cnf}
    \frac{\mathrm{d}\mathbf{x}(t)}{\mathrm{d}t} = f_\theta(\mathbf{x}(t))
\end{equation}
with boundary conditions $\mathbf{x}(t=0) \in \mathcal{D}$ and $\mathbf{x}(t=1) \sim q(\mathbf{x}')$. The continuous limit of repeatedly applying the change of measure identity yields the probability flow:
\begin{equation}
    \label{eq:probability-flow}
    \frac{\partial p(\mathbf{x}, t)}{\partial t} = \exp \left[ -\mathrm{tr}\;  J_{f_\theta}(\mathbf{x}(t))\right],
\end{equation}
which depends on the trace of the Jacobian -- a computation that scales linearly with $d$. Similar to normalizable architectures, a neural network $f_\theta$ parameterizes the drift (right-hand side of Eq. \ref{eq:cnf}) that minimizes the free energy difference between $\mathcal{D}$ and samples generated by $f_\theta$. 

Once parameterized, the change in probability is obtained by integrating the divergence of the probability flow over the generative trajectory, i.e.
\begin{equation}
    \label{eq:cnf-free-energy}
    \begin{aligned}
        \log p(\mathbf{x}(0)) &= \log q(\mathbf{x}(1)) \\
        &\quad - \int_0^1 \nabla \cdot 
        \left[ \mathrm{tr}\; J_{f_\theta}(\mathbf{x}(t)) \right] dt,
    \end{aligned}
\end{equation}
where the divergence can be approximated by the Hutchinson trace estimator\cite{hutchinson1989stochastic}.

Neural ODEs yield a change of coordinates that smoothly deforms $p(\mathbf{x})$ into $q(\mathbf{x}')$, and the neural ODE can be simulated forward or reverse in time to transport samples between the prior and target distributions and compute free energy differences. However, they are potentially difficult and expensive to parameterize since Eq. \ref{eq:cnf} must be simulated and backpropagation must carried out through the simulated trajectory\cite{CNF}.

\subsection{Diffusion Models}
\label{sec:sgm}
Diffusion models frame generative modeling in terms of a transport equation with an ODE solution that interpolates between $p(\mathbf{x})$ and $q(\mathbf{x}')$ \cite{nonequilibrium-learning,ddpm,sgm-sde,FPE-probability-flow}. The transport equation takes the form of a Fokker-Planck equation
\begin{equation}
    \label{eq:fokker-planck-equation}
    \frac{\partial p(\mathbf{x},t)}{\partial t} = -\lambda(t) \nabla \cdot \left[h(\mathbf{x},t)p(\mathbf{x},t)\right]
\end{equation}
describing the diffusion of a probability density under the influence of a vector-field $h(\mathbf{x},t)$ \cite{FPE-probability-flow}. 
The vector-field in Eq. \ref{eq:fokker-planck-equation} is chosen to result in a linear drift
\begin{equation}
    \label{eq:drift}
    h(\mathbf{x},t) = \mathbf{x} - \nabla \log p(\mathbf{x},t),
\end{equation}
so that the diffusion has the effect of transporting an arbitrary initial density $p(\mathbf{x})$ towards a Gaussian distribution $q(\mathbf{x}')$. The convergence is asymptotic, so a time-dilation factor $\lambda(t)$ is introduced to ensure that Eq. \ref{eq:fokker-planck-equation} is sufficiently converged at $t=1$.

The diffusion in Eq. \ref{eq:fokker-planck-equation} can equivalently be expressed as a stochastic differential equation (SDE) that transports samples from $p(\mathbf{x})$ to $q(\mathbf{x}')$:
\begin{equation}
    \label{eq:sde-fwd}
    \mathrm{d}\mathbf{x} = -\lambda(t) \mathbf{x} \mathrm{d}t + \sqrt{2 \lambda(t)}\mathrm{d}\mathbf{B}_t,
\end{equation}
where $\mathbf{B}_t$ is a Brownian motion. The linear drift allows for simulation free evaluation of the SDE, since the path distribution $p(\mathbf{x},t)$ originating from any $\mathbf{x}(t=0)$ has closed form \cite{sgm-sde,nonequilibrium-learning}.

A continuous-time generative model must be both (i) reversible to transport samples from $q(\mathbf{x}')$ to those of $p(\mathbf{x})$ and (ii) invertible to guarantee that the change of measure identity may be applied. Indeed, the diffusion equation is time-reversible under the change of variable $\tau = 1 - t$, and, remarkably, the time-reverse of the SDE in Eq. \ref{eq:sde-fwd} is
\begin{equation}
    \label{eq:sde-bck}
    \begin{aligned}
        \mathrm{d}\mathbf{x} =& -\lambda(\tau) \left[ \mathbf{x} + \nabla \log p(\mathbf{x},\tau) \right] \mathrm{d}\tau \\
        &+ \sqrt{2 \lambda(\tau)} \, \mathrm{d}\mathbf{B}_\tau.
    \end{aligned}
\end{equation}
The invertibility condition is satisfied when the variance of $\mathbf{B}_\tau$ is zero, with the resulting dynamics being described by the probability flow ODE:
\begin{equation}
    \label{eq:pf-ode}
    \frac{\mathrm{d}\mathbf{x}}{\mathrm{d}\tau} = -\lambda(\tau) \left[ \mathbf{x} + \nabla \log p(\mathbf{x},\tau) \right].
\end{equation}

The only unknown quantity in Eqs. \ref{eq:sde-bck} and \ref{eq:pf-ode} is $\nabla \log p(\mathbf{x},t)$ -- the {\it score}, which must be estimated \cite{hyvarinen-score}. Score-based models use a neural network $\mathbf{s}_\theta(\mathbf{x},t)$ to approximate $\nabla \log p(\mathbf{x},t)$ from realizations of Eq. \ref{eq:sde-fwd}. If the score estimate is sufficiently accurate, then equations \ref{eq:sde-bck} or \ref{eq:pf-ode} can be simulated to transport samples from $q(\mathbf{x}')$ to $p(\mathbf{x})$ and Eq. \ref{eq:cnf-free-energy} can be used to compute the free energy difference from the drift of the probability flow ODE.

\subsection{Schr\"odinger Bridges}
\label{sec:sbm}

One may further desire a transport process capable of mapping samples between arbitrary densities. Obtaining such a process amounts to solving the {\it Schr\"odinger Bridge} (SB) problem \cite{schrodinger-1932,leonard-2014,debortoli-2021,chen-2021}. The SB problem seeks to obtain the path distribution bridging distributions $p(\mathbf{x})$ and $q(\mathbf{x}')$ that minimizes the KL divergence to a reference path distribution. \citet{follmer1988} constructs the solution using diffusive dynamics and optimal transport: the optimal path reweights a reference Brownian path distribution with an entropically regularized optimal transport plan between $p(\mathbf{x})$ and $q(\mathbf{x}')$\cite{leonard2013survey}. Two related lines of work -- Stochastic Interpolants \cite{stochastic-interpolants,nfs-with-si,si-with-couplings} and Flow Matching \cite{flow-matching,flow-matching-ot,sb-flow-matching} -- have been developed to approximate the SB solution numerically using neural ODEs and SDEs.

\section{Experiments}
\label{sec:experiments}

Having examined the different flavors of probabilistic generative models, we turn now to numerical experiments to compare performance across two separate datasets. NS is an example of a normalizable architecture (Section \ref{sec:normalizing-flows}), CFM is a continuous flow model (Section \ref{sec:neural-ode}) which solves a user-selected bridge problem (Section \ref{sec:sbm}), in this case obeying the optimal transport solution between the target and prior distributions, and DDPM represents the broad diffusion model class (Section \ref{sec:sgm}). To compare the training, sampling, and accuracy of the NS, CFM, and DDPM models, we perform experiments on two datasets - a Gaussian mixture model in spaces of varying dimensionality and the dihedral torsion angle distribution associated with an Aib\textsubscript{9} molecular dynamics simulation in water (see Section \ref{sec:data-details}:Appendix for GMM data generation procedure and Aib\textsubscript{9} simulation details). We collected information on all configurational coordinates for the Gaussian mixture model, while for Aib\textsubscript{9} we collected information on all 9 $\{\Phi, \Psi\}$ pairs of dihedral angles. To quantify the accuracy of a given model, we begin by performing principal component analysis on the training data. After conducting training, generated samples are projected along the two principal components of the training data. This data projection is binned, and KL divergence is computed between vectors enumerating the counts in the corresponding bins.

\begin{figure}[hb]
    \centering
    \includegraphics[width=0.4\textwidth]{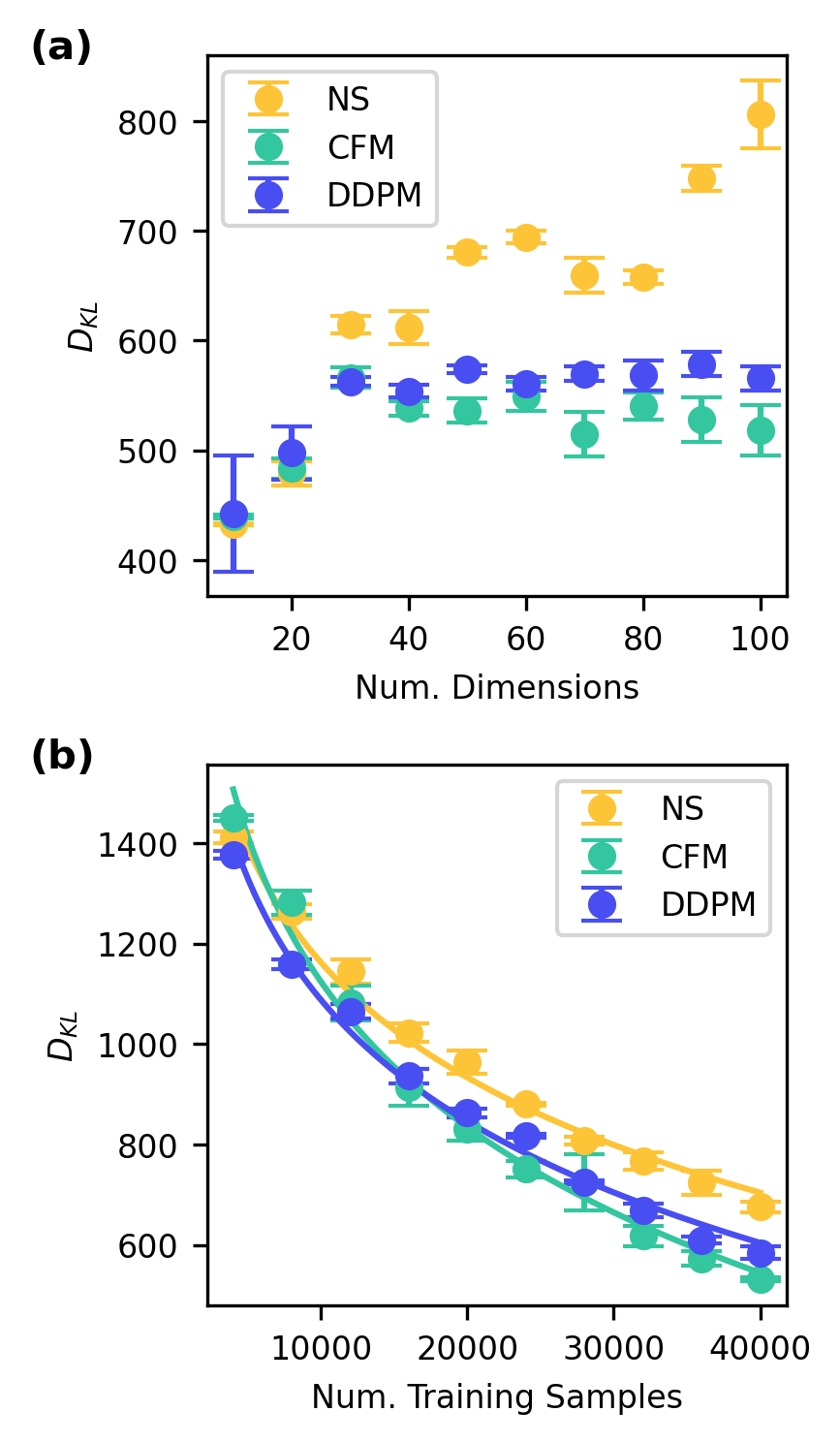}
    \caption{\textbf{Model accuracy results for the 4-modal Gaussian mixture dataset.} a) KL divergence ($D_{KL}$) comparison with analytical benchmark as a function of dimensionality. b) KL divergence comparison as a function of training dataset size at a fixed dimensionality of 50.}
    \label{fig:panel gmm main}
\end{figure}

In addition to measuring the accuracy, we also report the speed of sample generation and the number of learnable parameters, a proxy for model complexity. For each model, at least one hyperparameter controlling the neural network's depth or width was optimized. The complete set of experiments performed for each dataset is as follows:

\subsection*{Gaussian Mixture Experiments}
    \begin{itemize}
        \item Measure $D_{KL}$ for varying data dimensionality at a fixed number of modes and amount of training data (Figure \ref{fig:panel gmm main}a).
        \item Measure $D_{KL}$ for varying amounts of training data at a fixed number of modes and dimensionality (Figure \ref{fig:panel gmm main}b).
        \item Measure $D_{KL}$ for varying free energy difference between modes at a fixed dimensionality, number of modes, and amount of training data (Figure \ref{fig:panel asym}).
        \item Measure sample generation time for varying data dimensionality at a fixed number of modes and amount of training data (Figure \ref{fig:panel gmm si}a).
        \item Measure the number of neural network parameters for varying data dimensionality at a fixed number of modes and amount of training data (Figure \ref{fig:panel gmm si}b).
    \end{itemize}
    
\subsection*{Aib\textsubscript{9} Experiments}
    \begin{itemize}
        \item Measure $D_{KL}$ for the distribution of $\{\Phi, \Psi \}_i$ for each residue $i$ (Figure \ref{fig:panel aib9}a).
        \item Measure $D_{KL}$ along $\{\Phi, \Psi \}_5$ for varying amount of training data (Figure \ref{fig:panel aib9}b).
    \end{itemize}

\subsection{Gaussian Mixture}
\label{sec:gmm}

The first numerical experiment we perform concerns a Gaussian mixture, a superposition of samples drawn from independent Gaussian distributions centered at different locations, with four modes but dimensionality varying from 10 to 100. After generating the data for the Gaussian mixture, we hold back $10\%$ of the data as a test set and allow the models to train for an equal amount of time. Figure \ref{fig:panel gmm main}a shows that for datasets of dimensionality 40 or greater, we find a lower KL divergence, and thus higher accuracy, for CFM compared to DDPM and especially NS. The performance of CFM and DDPM seems to remain stable with no upward trend as dimensionality increases beyond 40, unlike NS, which rises sharply at very high dimensionality.

Our next GMM experiment considers varying the amount of training data the model sees and measuring model accuracy. In this test, the dimensionality is fixed at 50, and we consider a Gaussian mixture with four modes. Figure \ref{fig:panel gmm main}b shows that KL divergence as a function of training dataset size varies logarithmically for all three models, and DDPM fares better than NS and CFM for low amounts of training data.

The third experiment performed on the Gaussian mixture dataset examines the ability of each model to reproduce free energy differences found in training data. To this aim, we designed a training dataset with a 50-dimensional bimodal Gaussian mixture where the free energy difference between the two modes varies. This construction differs from the 4-modal distribution of the other GMM experiments so that the free energy difference between the two modes can be isolated and examined. We can compute the free energy difference between the two modes by first establishing a boundary defining two domains corresponding to the two states in the PCA histogram space. To compute the free energy difference between the two modes, we first compute the partition function for each state using $Z_i = \sum_{j \in i} p_j$, where $i$ is a state label and $p_j$ represents the probability associated with histogram bin $j$. For both numerical stability and to exclude low-probability, high-free-energy bins, we impose a free-energy minimum cutoff of roughly 0.0374 kJ/mol from the corresponding energy minimum and only sum over bins meeting this criterion for either domain. The free energy difference is then $\Delta F = -{1 \over \beta} \mathrm{ln} (Z_1/Z_2)$, where the partition functions correspond to the two modes, and we have used $\beta=1$.

\begin{figure}[ht]
    \centering
    \includegraphics[width=0.4\textwidth]{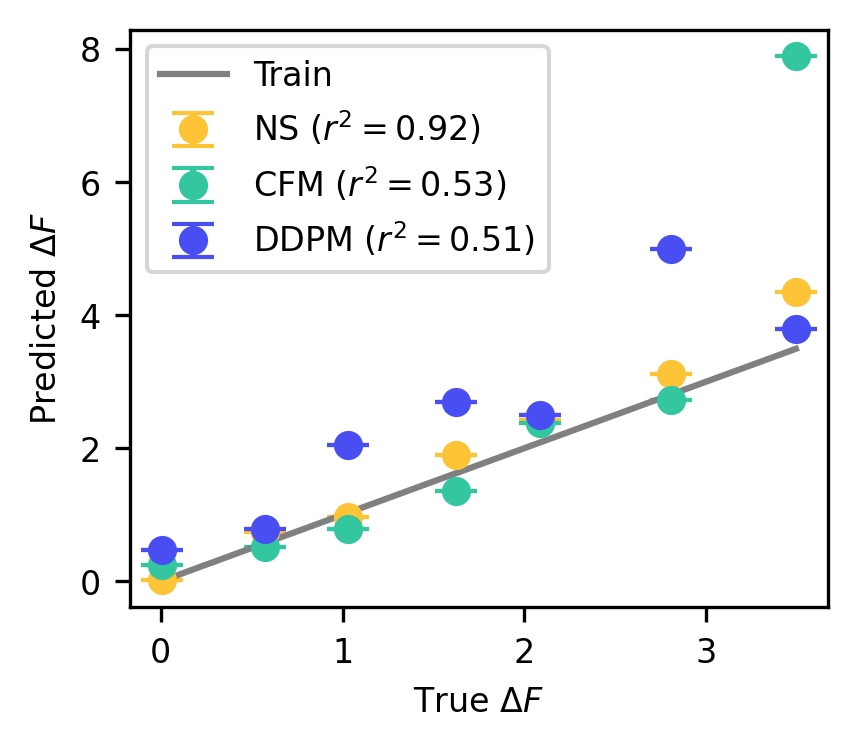}
    \caption{\textbf{Free energy difference estimation accuracy on asymmetric bimodal distributions.} A free energy difference of zero represents two equal Gaussian modes, while higher free energy indicates a higher level of asymmetry. $r^2$ is computed with the residuals of each model from the plotted line indicating training free energy difference. We impose a free energy cutoff of 0.0374 kJ/mol and note the data dimensionality is fixed at 50.}
    \label{fig:panel asym}
\end{figure}

Figure \ref{fig:panel asym} shows the accuracy results and the coefficient of determination $r^2$ for each model compared to the training free energy difference. We conclude that NS reproduces the training free energy differences the most faithfully, followed by CFM and DDPM. It is useful to note that the least accurate model by KL divergence in our principal GMM dimensionality test outperforms the other two in this case.

\begin{figure}[ht]
    \centering
    \includegraphics[width=0.4\textwidth]{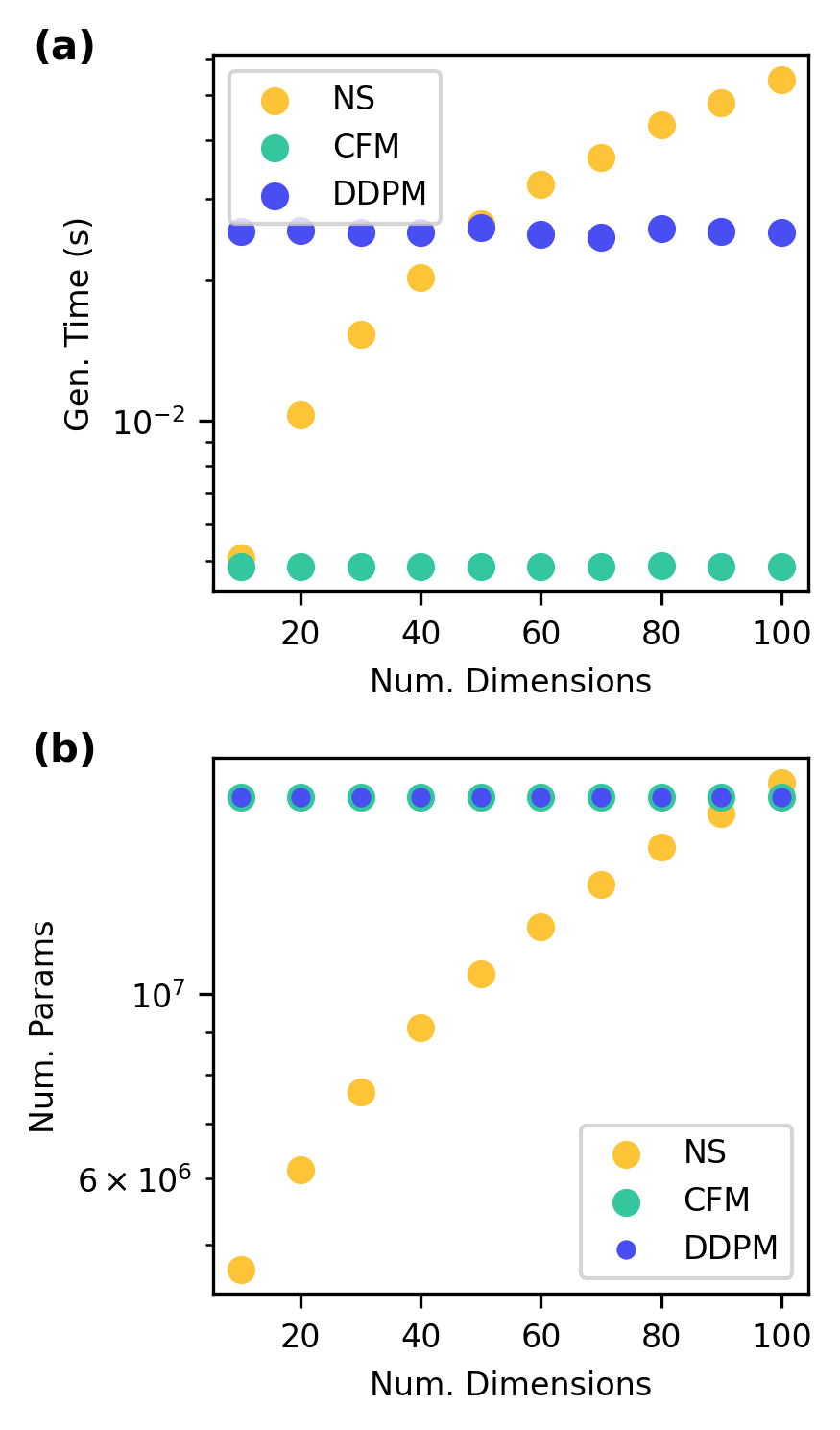}
    \caption{\textbf{Speed and model size results for the 4-modal Gaussian mixture dataset.} a) Single sample generation time comparison as a function of dimensionality. b) Model capacity comparison measured by total number of learnable parameters as a function of dimensionality.}
    \label{fig:panel gmm si}
\end{figure}

Our final two GMM experiments concern sampling speed and model capacity, both measured as dimensionality varies for a Gaussian mixture with four modes. As shown in Figure \ref{fig:panel gmm si}a, CFM displays much faster inference than DDPM or NS due to its inexpensive calls to an ODE solver rather than the reverse simulation of an SDE or propagation of samples through increasingly complex splines involving high-dimensional algebraic operations. Figure \ref{fig:panel gmm si}b shows the results of the model capacity measurement. CFM and DDPM employ the same predictive neural network, so their model capacity is equal. In contrast, the NS network is initially less expensive but increases rapidly in size as dimensionality climbs.

\subsection{Aib\textsubscript{9} Dihedral Torsion Angles}
\label{sec:aib9}

The previous example has a limit of four modes in the underlying probability distribution. More often than not, molecular systems tend to have a very large number of modes corresponding to different metastable states. To mimic this situation, we move to a more complex system. Aib\textsubscript{9} is a synthetic peptide used as a model system because of its clear chirality transitions between left and right-handed forms. We simulate a molecular dynamics trajectory of an Aib\textsubscript{9} molecule and record all of the 9 $\{ \Phi,\Psi \}_i$ dihedral torsion angle pairs (18 angles in total) as a function of simulation time (see Section \ref{sec:data-details}:Appendix for details of the molecular dynamics data generation procedure). In this experiment, we input a collection of 18-dimensional vectors of $\{ \Phi,\Psi \}_i$ values as training data and train the models to generate samples from this target distribution. KL divergence is then computed for each residue individually by isolating the corresponding $\{ \Phi,\Psi \}_i$ pair and performing principal component analysis between the held-back test set and generated samples.

The `exterior' residues closest to the end of the peptide chain, residues 1 and 9, for example, are more flexible and thus more rapidly transition between left and right-handed orientations, sampling more of the transition states and other high-energy regions. These residues thus have a more complex distribution than the middle residues, which exhibit slower transitions \cite{aib9-residues}. This behavior can be seen in the U-shaped curve of KL divergence scores, indicating the higher difficulty the models face in generating the exterior residues.

\begin{figure*}[ht]
    \centering
    \includegraphics[width=1\textwidth]{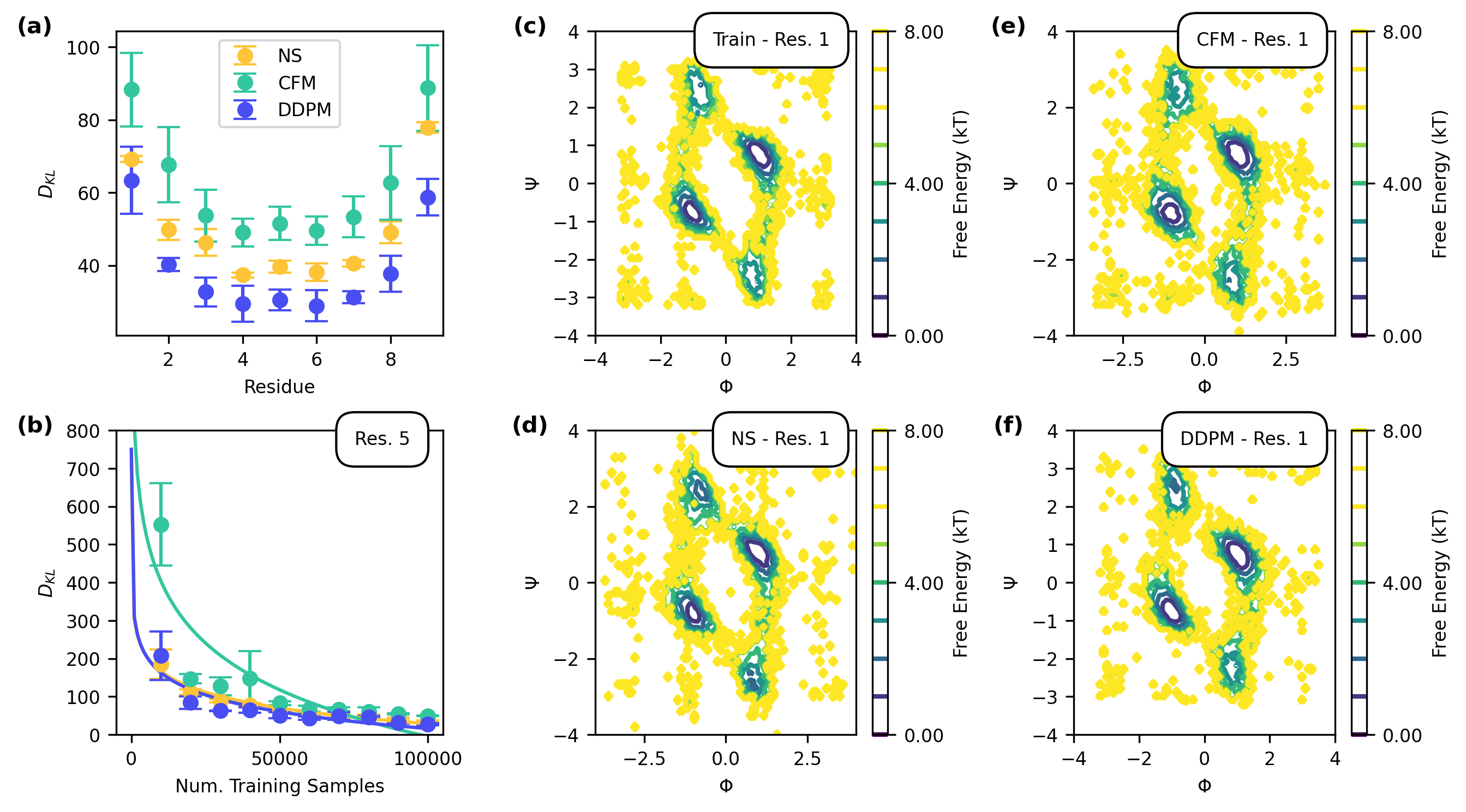}
    \caption{\textbf{Model accuracy results and generated data for the Aib\textsubscript{9} peptide.} a) KLD performance comparison as a function of residue index for the complete Aib\textsubscript{9} torsion angle dataset. b) KLD performance comparison for the three models and Gaussian baseline fit as a function of training dataset size for the Aib\textsubscript{9} torsion angle data distribution at residue 5. c) d) e) f) $\{ \Phi,\Psi \}$ show free energy contour plots for torsion angle distributions at residue 1 for training data and generated data for NS, CFM, and DDPM, respectively.}
    \label{fig:panel aib9}
\end{figure*}

Training is performed exactly as in the Gaussian mixture experiment, and accuracy is measured via the KL divergence at individual residues via the procedure described above. Correspondingly, training is done in an 18-dimensional space, and measurement is done in a 2-dimensional space (principal components corresponding to $\{ \Phi,\Psi \}_i$ torsion angles of a given residue). Figure \ref{fig:panel aib9}a shows the accuracy results as a function of residue index. We observe the highest accuracy from DDPM, with CFM performing the least well universally.

We show the results of varying training dataset sizes in Figure \ref{fig:panel aib9}b. In this experiment, the residue considered was fixed as residue 5. DDPM and NS perform comparably and show better accuracy than CFM for all dataset sizes. $\{ \Phi,\Psi \}_4$ plots at residue 1 are shown in Figure \ref{fig:panel aib9}c-f respectively for training data and generated data for NS, CFM, and DDPM.

\section{Conclusion}
\label{sec:conclusion}

In this work, we explored qualitatively and quantitatively how three different classes of probabilistic generative models perform on datasets with tunable, varied complexity. We considered three classes of models, namely Neural Spline Flows (NS) models, Conditional Flow Matching (CFM) models, and Denoising Diffusion Probabilistic Models (DDPM). This selection of models is by no means exhaustive, and recently introduced architectures such as Rectified Flow \cite{rect-flow}, Latent Diffusion Models \cite{rombach2022high}, and newly improved diffusion architectures \cite{karras2022elucidating} stand out as prime candidates for future testing with these benchmark datasets. For our chosen models, we performed experiments on a Gaussian mixture model and molecular dynamics simulations of a 9-residue synthetic peptide undergoing chirality transitions in water. 
After introducing each class of generative architecture, exploring strengths and weaknesses, and examining the results of our model comparison experiments on the Gaussian mixture and Aib\textsubscript{9} torsion angle datasets, we may now draw some conclusions about the relative cases in which each model is the optimal choice. CFM outperforms other models for high-dimensional datasets of limited complexity, such as the Gaussian mixture model, and exhibits the fastest inference. For lower-dimensional datasets of high complexity, such as the  Aib\textsubscript{9} torsion angle dataset, DDPM is the most accurate. For the free energy difference estimation task, NS most accurately reproduces asymmetry between modes in the training data. We hope these conclusions will help guide the selection of models for a given task depending on the characteristics of the training data, including dimensionality, complexity, and asymmetry, and how sensitive the generative problem is to accuracy, cost, and speed. We also expect that the systematic curation of datasets with quantified complexity will be helpful for future methods developed for probabilistic generative modeling and beyond.

\section{Data and Code Availability}
\label{sec:availability}

Code to train and sample from all models, as well as perform the experiments, is available at \url{https://github.com/tiwarylab/model-comparison}. Code to generate the GMM datasets is available at the above link. All datasets used are available on
\href{https://zenodo.org/records/14143082?token=eyJhbGciOiJIUzUxMiJ9.eyJpZCI6IjAyYmYzODhlLWE2ZjYtNDA4NS1iNDhlLTJlNzZmMzcyNzMwZCIsImRhdGEiOnt9LCJyYW5kb20iOiI0YTE3NTE3N2Y4MThkODg0YTY4NTI4OWExMGE3NmNmNiJ9.HcFgvUV0sK8EhJm0Ow8cFn-56q8rGuSWj_LBQIcpzMZ_mAySqnJ4pJeJubxw_3Dtl2chUoHAGOaxgaRFyZRLWg}{Zenodo}. Additionally, while the code to run molecular dynamics simulations for Aib\textsubscript{9} is available at the above address, a more comprehensive package is available at \url{https://github.com/shams-mehdi/aib9_openmm}.

\section{Acknowledgements}
\label{sec:acknowledgments}

We thank the developers of the \texttt{normalizing-flows} and \texttt{TorchCFM} GitHub repositories for their implementations used in this work. This research was entirely supported by the US Department of Energy, Office of Science, Basic Energy Sciences, CPIMS Program, under Award DE-SC0021009. We thank UMD HPC's Zaratan and NSF ACCESS (project CHE180027P) for computational resources. P.T. is an investigator at the University of Maryland-Institute for Health Computing, which is supported by funding from Montgomery County, Maryland and The University of Maryland Strategic Partnership: MPowering the State, a formal collaboration between the University of Maryland, College Park and the University of Maryland, Baltimore. \newline

\section{Appendix}
\label{sec:appendix}

\subsection{Data Details}
\label{sec:data-details}

Details of the Aib\textsubscript{9} molecular dynamics simulation are provided in Table 1.

\begin{table}[h!]
\centering
\begin{tabular}{|c|c|}
\hline
\textbf{Parameter} & \textbf{Value} \\
\hline
Simulation engine & OpenMM \\
Temperature & 450 K \\
Water model & TIP3 \\
Integration step & 2 fs \\
Energy minimization & True \\
NVT equilibration & 1 ns \\
NPT equilibration & 1 ns \\
Production run & 200 ns \\
\hline
\end{tabular}
\vspace{10pt} 
\caption{Aib\textsubscript{9} MD parameters}
\end{table}

The Gaussian mixture model data was generated via \texttt{torch.distributions}. This package enables sampling from Gaussian distributions of arbitrary dimensionality with given mean vectors and covariance matrices. In our usage, covariance matrices are the identity matrix, and mean vectors are drawn randomly. To generate multimodal Gaussian mixtures, we draw samples from independent Gaussian distributions and combine them. The ratio of samples from one distribution versus another can be varied, allowing a controllable degree of asymmetry for our free energy difference estimation experiment.

As noted in Section \ref{sec:availability}, all code to perform data generation (GMM and Aib\textsubscript{9} datasets) is available at \url{https://github.com/tiwarylab/model-comparison}, and the datasets are additionally located on \href{https://zenodo.org/records/14143082?token=eyJhbGciOiJIUzUxMiJ9.eyJpZCI6IjAyYmYzODhlLWE2ZjYtNDA4NS1iNDhlLTJlNzZmMzcyNzMwZCIsImRhdGEiOnt9LCJyYW5kb20iOiI0YTE3NTE3N2Y4MThkODg0YTY4NTI4OWExMGE3NmNmNiJ9.HcFgvUV0sK8EhJm0Ow8cFn-56q8rGuSWj_LBQIcpzMZ_mAySqnJ4pJeJubxw_3Dtl2chUoHAGOaxgaRFyZRLWg}{Zenodo}.

\begin{figure*}[b]
    \centering
    \includegraphics[width=.9\textwidth]{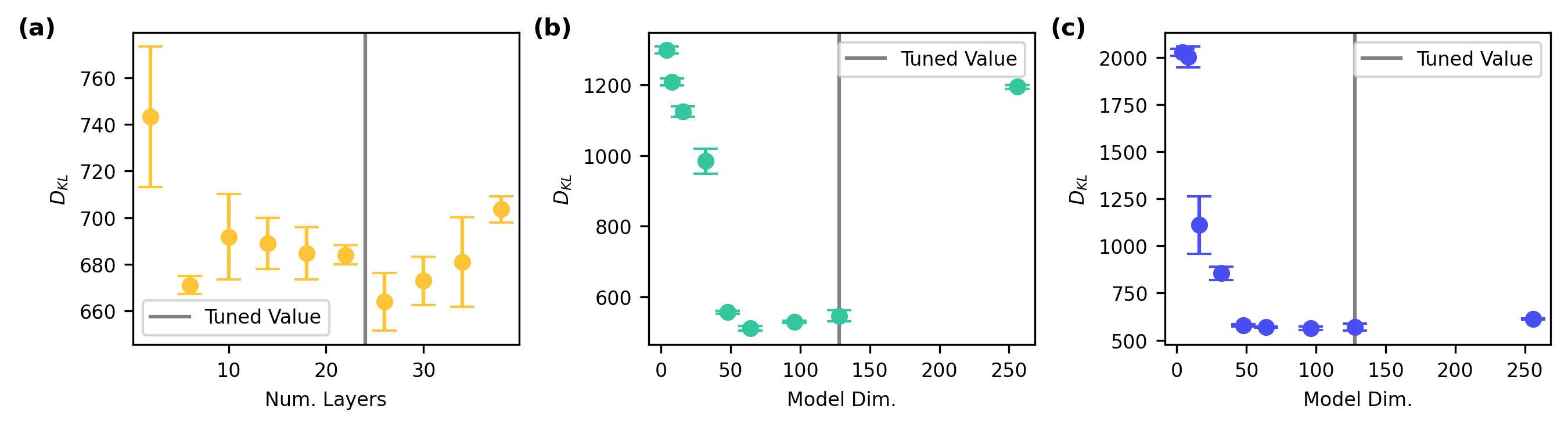}
    \caption{\textbf{Hyperparameter tuning for each model.} a) Tuning of the `layers' parameter of NS. b) c) Tuning of the `model dimension' parameter of CFM and DDPM respectively.} 
    \label{fig:panel tuning}
\end{figure*}

\subsection{Additional Figures}
\label{sec:addlfig}

Figure \ref{fig:panel tuning} shows the hyperparameter tuning results for the three generative models. The `layers' parameter of NS corresponds to the depth of the network, whereas the `model dimension' parameter of CFM and DDPM corresponds to the width and, implicitly, the depth of the network.

\end{document}